\documentclass[conference]{IEEEtran}

\usepackage{color}
\usepackage{algorithm}
\usepackage{algpseudocode}
\usepackage{pseudocode}
\usepackage{graphicx}
\usepackage{amsmath}
\usepackage{csquotes}
\usepackage{multirow}
\usepackage{tikz}
\usepackage{subcaption}
\usepackage{etoolbox}
\usepackage{amssymb,amsthm,enumitem}
\usepackage{microtype}
\usepackage{booktabs} 
\usepackage{soul}
\usepackage{hyperref}
\usepackage{multirow}
\usepackage{caption}
\usepackage[bottom]{footmisc}
\usepackage{cite}
\usepackage[flushleft]{threeparttable}

\newcommand{\sys}{CodeX} 
\newcommand{\mystyle}{\textit}
\graphicspath{{figures/}} 
\newcommand{\mli}[1]{\mathit{#1}}

\usepackage{pifont}

\begin{document}

\title{\bf \Large \sys{}: Bit-Flexible Encoding for Streaming-based FPGA Acceleration of DNNs
\vspace{-1em}}


\author{\IEEEauthorblockN{Mohammad Samragh,
Mojan Javaheripi,
Farinaz Koushanfar}
\IEEEauthorblockA{Department of Electrical and Computer Engineering,
University of California San Diego\\ \{msamragh, mojan, farinaz\}@ucsd.edu}}


\maketitle
\begin{abstract}
This paper proposes \sys{}, an end-to-end framework that facilitates encoding, bitwidth customization, fine-tuning, and implementation of neural networks on FPGA platforms. \sys{} incorporates nonlinear encoding to the computation flow of neural networks to save memory. The encoded features demand significantly lower storage compared to the raw full-precision activation values; therefore, the execution flow of \sys{} hardware engine is completely performed within the FPGA using on-chip streaming buffers with no access to the off-chip DRAM. We further propose a fully-automated algorithm inspired by reinforcement learning which determines the customized encoding bitwidth across network layers. \sys{} full-stack framework comprises of a compiler which takes a high-level Python description of an arbitrary neural network architecture. The compiler then instantiates the corresponding elements from \sys{} Hardware library for FPGA implementation. Proof-of-concept evaluations on MNIST, SVHN, and CIFAR-10 datasets demonstrate an average of $\mathbf{4.65\times}$ throughput improvement compared to stand-alone weight encoding. We further compare \sys{} with six existing full-precision DNN accelerators on ImageNet, showing an average of $\mathbf{3.6\times}$ and $\mathbf{2.54\times}$ improvement in throughput and performance-per-watt, respectively.  
\end{abstract}

\IEEEpeerreviewmaketitle

\vspace{-0.2em}
\section{Introduction}
Deep Neural Networks (DNNs) are being widely developed for various machine learning applications, many of which are required to run on embedded devices. In the realm of embedded DNNs, just-in-time execution under severe power limitations is hard to satisfy~\cite{li2017deeprebirth,zhang2017shufflenet,howard2017mobilenets}. Contemporary research has focused on FPGA-based acceleration of DNNs~\cite{zhang2015optimizing,ovtcharov2015accelerating,suda2016throughput,ghasemzadeh2018rebnet,sharma2016high,samragh2017customizing,umuroglu2017finn,prost2017scalable,shea2018scalenet}. However, FPGAs are inherently limited in terms of on-chip memory capacity. Thus, the high-storage requirement of DNN models hinders an efficient and low power execution on FPGAs.

To reduce the computational complexity and memory requirement of DNNs, several pre-processing algorithms have been proposed. The existing methods generally convert conventional DNNs into compact representations that are better suited for execution on embedded devices. Examples of such compacting methods include quantization~\cite{hubara2016quantized}, binarization~\cite{ghasemzadeh2018rebnet,umuroglu2017finn}, tensor decomposition~\cite{kim2015compression,zhang2015efficient,nazemi2018hardware}, parameter pruning~\cite{liu2015sparse,wen2016learning}, and compression with nonlinear encoding~\cite{han2015deep,samragh2017customizing}. Higher compression rate might not always translate to the hardware performance optimization as the platform constraints could interfere with the intended compaction methodology \cite{yang2018netadapt}. 



This paper specifically focuses on nonlinear encoding and provides solutions to tackle the challenges associated with optimizing physical performance. Encoding network parameters is rather beneficial as it reduces memory footprint, i.e., the main source of delay and power consumption in FPGA accelerators. To devise a practical solution for implementing encoded DNNs, we simultaneously identify and address four critical issues.



\begin{figure}
    \centering
    \includegraphics[width=0.7\columnwidth]{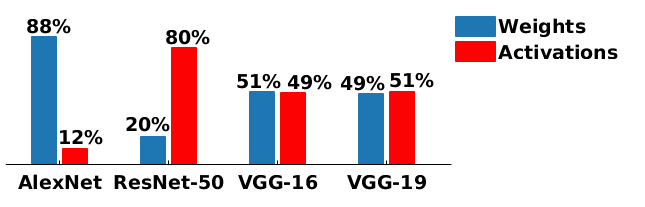}
    \caption{Relative memory footprint of weights and activations for various DNN architectures, evaluated on 10 samples.}
    \label{fig:relative_mem}
    \vspace{-1.5em}
\end{figure}

First, DNN memory footprint is imposed by either weights or feature-maps. Figure~\ref{fig:relative_mem} shows the relative memory requirements in several popular DNN models. As can be seen, the memory footprint of activations is notable; however, contemporary research mainly targets the (static) DNN weights for nonlinear quantization~\cite{samragh2017customizing,han2015deep,chen2015compressing}. Developing online mechanisms for activation encoding can significantly reduce the memory footprint of DNN models. Second, nonlinear quantization destabilizes DNN training by adding non-differentiable elements to the model. Therefore, novel computation routines must be developed to approximate gradients for DNN fine-tuning. Third, specifying the encoding bitwidth across all DNN layers by handcrafted try-and-error is exhaustive and generally sub-optimal. Hence, automated and intelligent solutions are highly preferable. Finally, designing accelerators that are customized per application/hardware is cumbersome. As such, easy-to-use tools are needed to ensure low, non-recurring engineering cost. 


To tackle the aforementioned challenges, we introduce \sys{}, a unified framework that facilitates encoding, training, bitwidth customization, and automated implementation of encoded DNNs on FPGA platforms.
The distinguished contributions of this paper are listed as follows: 
\begin{itemize}
\setlength\itemsep{0.2em}
    \item Introducing a novel methodology for (online) encoding of DNN activations. We establish the gradient computation routines required to fine-tune the non-differentiable encoded DNNs, allowing fast restoration of DNN accuracy.
    \item Introducing an automated algorithm for customizing per-layer encoding bitwidths. Inspired by reinforcement learning, we establish an action-reward-state system to find a bitwidth configuration that minimally affects DNN accuracy and maximally reduces memory footprint.
    \item Establishing a hardware library for bit-flexible implementation of customized encoded DNN layers. Activation encoding lowers memory footprint and facilitates the use of streaming buffers for inter-layer feature transmission.  
    \item Providing an API for fast and easy hardware implementation of encoded DNNs. Developers describe the DNN as high-level Python code which is then automatically converted to Vivado\_HLS (the high-level programming language for Xilinx FPGAs). The API is open-source and can be used by the community\footnote{\url{https://github.com/MohammadSamragh/CodeX}}. 
\end{itemize}
\section{Overview and insights}
\sys{} design flow is composed of an interlinked optimization scheme where algorithmic DNN compaction methods and hardware-level customization for the accelerator are performed in sync. In this section, we describe \sys{} insights in high-level and look at the main components of our framework.

\subsection{Streaming-based On-chip Execution}
\begin{figure}[h]
\centering
\includegraphics[width=\columnwidth]{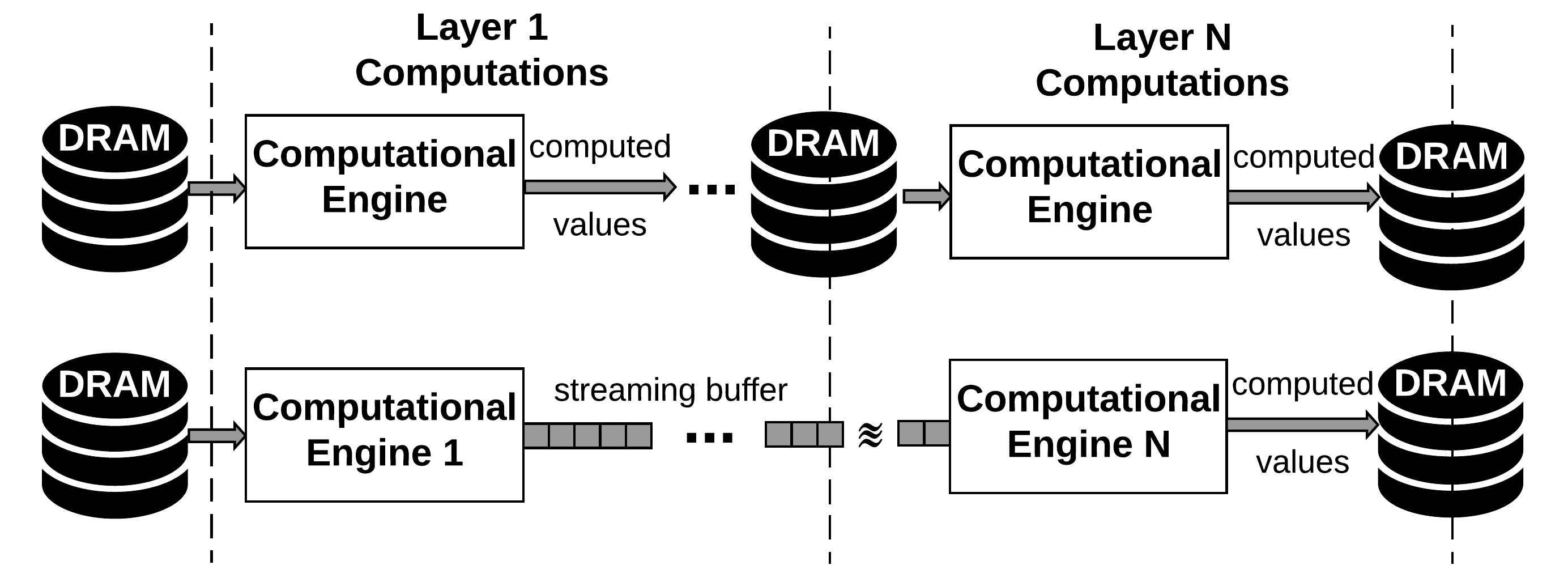}{\centering}
\vspace{-1.2em}
\caption{Workflow of traditional vs. streaming-based DNN inference. The top diagram illustrates the conventional approach where all resources are allocated to one computational engine and layer input/outputs are continuously read from and written to the off-chip memory. The bottom diagram presents a streaming-based methodology where communications with off-chip memory are limited to the first and last layer and each layer is allocated a different computational engine.}
\label{fig:stream_vs_static}
\end{figure}

Traditional DNN accelerators store the weights and activations (features) of layers in the off-chip DRAM since commodity FPGAs are often limited in terms of on-chip memory capacity. Figure~\ref{fig:stream_vs_static}-top demonstrates the computation flow of DNNs in such settings. Alternatively, the weights and computed activations could be stored and accessed within the FPGA design using streaming buffers as depicted in Figure~\ref{fig:stream_vs_static}-bottom. The benefits of the latter approach are three-fold. First, it avoids the power-hungry and high-latency access to off-chip DRAM. Second, the computation engines responsible for each DNN layer can be customized to comply with the pertinent layer. Finally, the streaming buffers allow pipelining the computation engines to increase throughput. 

Although on-chip execution of DNNs is beneficial in many aspects, the memory requirement of the weights and activations of DNN layers is often beyond the (limited) capacity of commodity FPGAs. To address this issue, \sys{} employs nonlinear quantization to reduce the memory footprint such that the weights/activations can be accommodated within FPGA block-RAMs. In the next section, we explain our nonlinear quantization methodology in high level.

\vspace{-1em}
\subsection{Memory Compression}
\begin{figure}[h]
    \centering
    \begin{subfigure}[b]{0.45\columnwidth}
            \includegraphics[width=0.95\columnwidth]{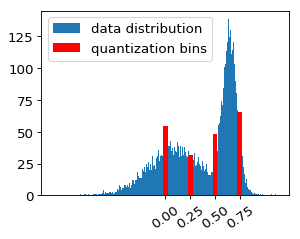}
            \caption{Fixed-point}
    \end{subfigure}
    ~
    \begin{subfigure}[b]{0.45\columnwidth}
        \includegraphics[width=0.95\columnwidth]{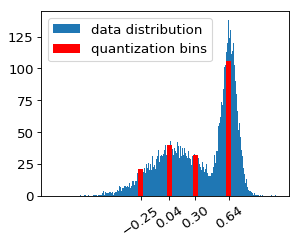}
        \caption{Nonlinear}
    \end{subfigure}
    \caption{Illustration of fixed-point and nonlinear quantization. In this example, there are 4 quantization bins and the approximations are represented with $\mathbf{Log_2(4)=2}$ bits. }
    \label{fig:linear-vs-nonlinear}
    \vspace{-1.5em}
\end{figure}

In quantization, a finite set of best representatives (a.k.a. bins) are selected and each real-valued number is approximated with the closest bin. Perhaps the most popular quantization is fixed-point approximation. Figure~\ref{fig:linear-vs-nonlinear}-a depicts the bins for an unsigned fixed-point quantization. In this setting, quantization bins are fixed to certain points (e.g., 0, 0.25, ...) regardless of data distribution. Alternatively, in nonlinear quantization, the bins are carefully selected to best represent the data. Figure~\ref{fig:linear-vs-nonlinear}-b shows the bins for such nonlinear quantization. In this example, both fixed-point and nonlinear quantizations require the same number of representation bits. However, the approximation error associated with the nonlinear scheme is drastically lower.  

\subsection{Global Flow}
\begin{figure}[ht]
    \centering
    \includegraphics[width=1\columnwidth]{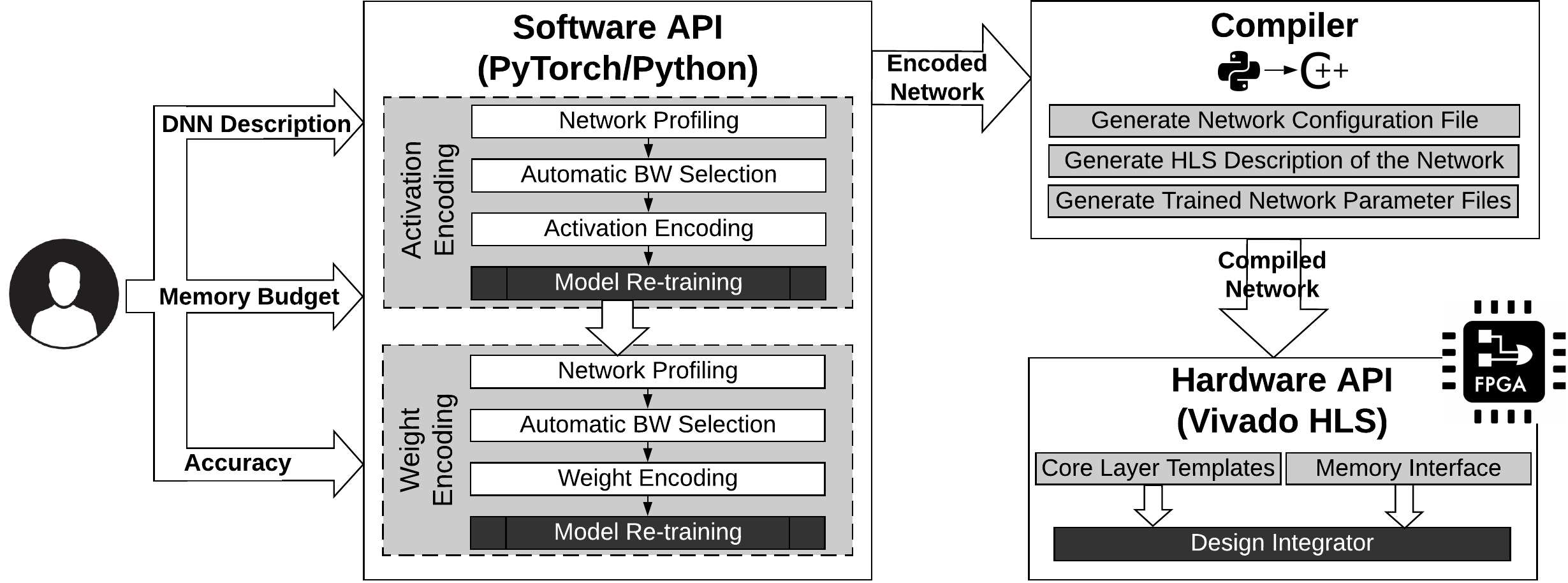}
    \caption{Global flow of \sys{} framework. The user provides a high-level Python description of a pre-trained DNN to the pre-processing module, which is responsible for weight/activation encoding, layer-specific bitwidth configuration, and model fine-tuning. The compiler converts the Python code into a hardware description. The hardware API provides a customized library for DNN synthesis and FPGA bitfile generation.}\label{fig:global_flow}
\end{figure}

Figure~\ref{fig:global_flow} presents the global flow of \sys{} framework consisting of an offline processing module, a compiler, and a hardware implementation library. 

\noindent{\bf Software API. } This module analyzes a given DNN (described in Python code) and applies nonlinear quantization to the layer weights/activations. \sys{} encoding scheme reduces memory footprint at the cost of a small reduction in classification accuracy. We devise an automated algorithm to determine the per-layer number of quantization bins for the weights/activations such that the memory footprint is maximally reduced and/or the accuracy is minimally affected. The software API also fine-tunes the quantized DNN to compensate for loss of accuracy. 

\noindent{\bf Hardware API. } The DNN hardware is described using Vivado\_HLS, which is a standard high-level-synthesis tool that enables faster development as well as portability. \sys{} provides a library of template functions that can realize various DNN functionalities. An arbitrary network architecture can be described by instantiating the corresponding templates. Each function has a set of configurable primitives such as the number of input/output neurons of the layer, the weights/activation bitwidths, and the parallelism factors for the layer execution. We will elaborate more on the hardware in Section~\ref{sec:hardware}.

\noindent{\bf Compiler. }To ensure ease-of-use and design automation, \sys{} is accompanied by a compiler that converts the high-level Python code for the DNN into a hardware description. \sys{} compiler produces a configuration file with the customized per-layer quantization bitwidths. The hardware description together with the configuration file enable instantiation of core layer modules. In addition, the compiler converts the encoded DNN parameters into a format ready for loading to the FPGA on-chip memory upon execution. 
\section{\sys{} Software Stack}\label{framework}

In this section we elaborate on the concepts utilized for nonlinear encoding of DNN parameters/activations. Section~\ref{sec:encoding} explains \sys{} weight/activation encoding method. Our customized gradient computations for training the encoded networks are formulated in Section~\ref{sec:training}. Finally, the automated bitwidth selection routine is explained in Section~\ref{sec:customization}.

\subsection{Encoding Scheme}\label{sec:encoding}



Our encoding scheme aims to estimate the parameters of a DNN layer with a subset of representatives referred to as the \mystyle{codebook}. In the rest of this section, we delineate \sys{} encoding method for DNN weights/activations.

\subsubsection{\textbf{Weight Encoding}}
\begin{figure}[h]
    \centering
    \includegraphics[width=0.95\columnwidth]{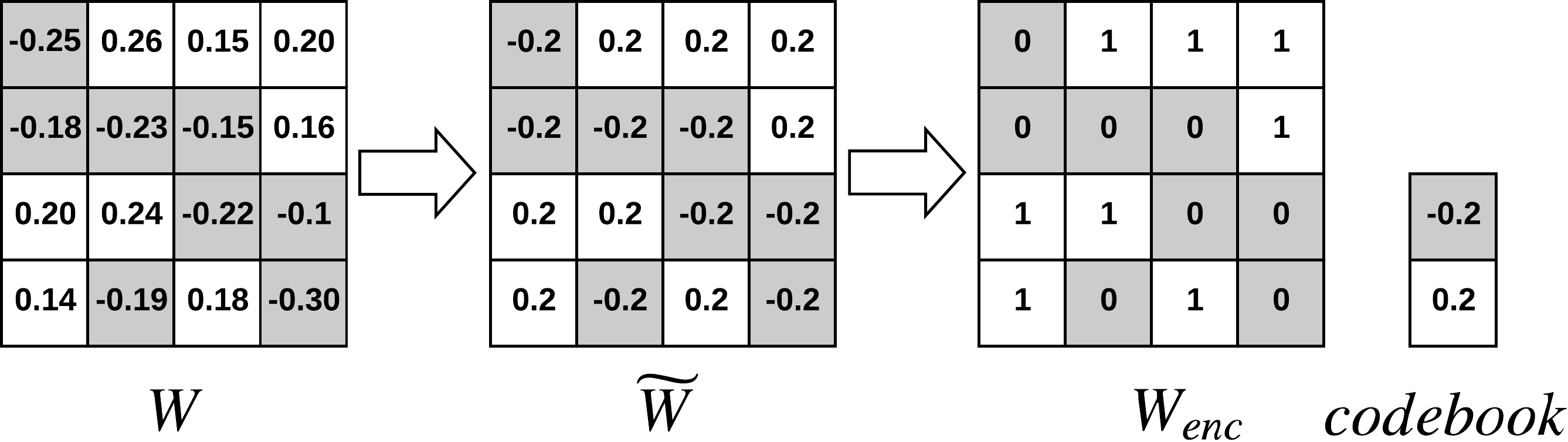}
    \vspace{-0.2em}
    \caption{Illustration of \sys{} weight encoding. left: original matrix $W$, middle: approximated matrix $\widetilde{W}$, right: encoded matrix $W_{enc}$ along with the codebook vector $c$.}
    \label{fig:encoding_w}
    \vspace{-0.5em}
\end{figure}

Let us denote the weight parameters in a certain DNN layer as $W$. In order to encode $W$, we first find an approximation $\widetilde{W}\approx W$ such that the elements of $\widetilde{W}$ are restricted to a finite set of real-values, $\vec{c}=\{c[1], \dots,c[K]\}$, i.e., the \textit{codebook}. Figure~\ref{fig:encoding_w} illustrates this approximation for a $4\times4$ matrix $W$ using a codebook of $K=2$ elements. We denote the encoded $\widetilde{W}$ as $W_{enc}$ where the elements are the index of the corresponding codebook value. 

To approximate $\widetilde{W}$ with a codebook of size $K$, authors of~\cite{han2015deep} suggest using the well-known K-means clustering algorithm~\cite{lloyd1982least}. While K-means can effectively solve the aforementioned problem for a fixed codebook size, specifying the codebook sizes in different layers of a neural network is a challenge yet to be solved. Specifically, different layers require different codebook sizes to capture the statistical properties of the pertinent parameters. To tackle this challenge, \sys{} proposes an automated bitwidth selection algorithm as explained in Section~\ref{sec:customization}.
Note that weight encoding is performed only once in an offline pre-processing step. The per-layer per-layer encoded weights and codebooks are then stored in binary files to be loaded in the FPGA memory.

\subsubsection{\textbf{Activation Encoding}}\label{sec:activation_encoding}
The encoding of activations is performed in two phases: (i)~offline phase performed in \sys{} software API, where the layer codebooks are generated using the K-means algorithm. (ii)~Online phase where each computed feature is encoded by the closest codebook value.


\noindent{\bf Offline K-means.} Algorithm 1 summarizes the process for computing DNN activation codebooks. First, a subsampled data set, $\vec{x}_1, \vec{x}_2, \dots, \vec{x}_N$, is used to generate the layer feature-maps, which we denote by $\vec{y}^{\ l}$. Next, $\vec{y}^{\ l}$ is flattened into an array, $\vec{a}^{\ l}$. Since layers with \mystyle{ReLU} nonlinearity produce many $0$-valued outputs, we only perform K-means on the non-zero elements of $\vec{a}^{\ l}$ which significantly reduces the runtime of the (offline) K-means clustering. The ($K^l-1$) cluster centers along with the appended $0$ value form the $l^{th}$ layer's codebook.
\begin{figure}[h]
    \centering
    \vspace{-1em}
    \includegraphics[width=0.85\columnwidth]{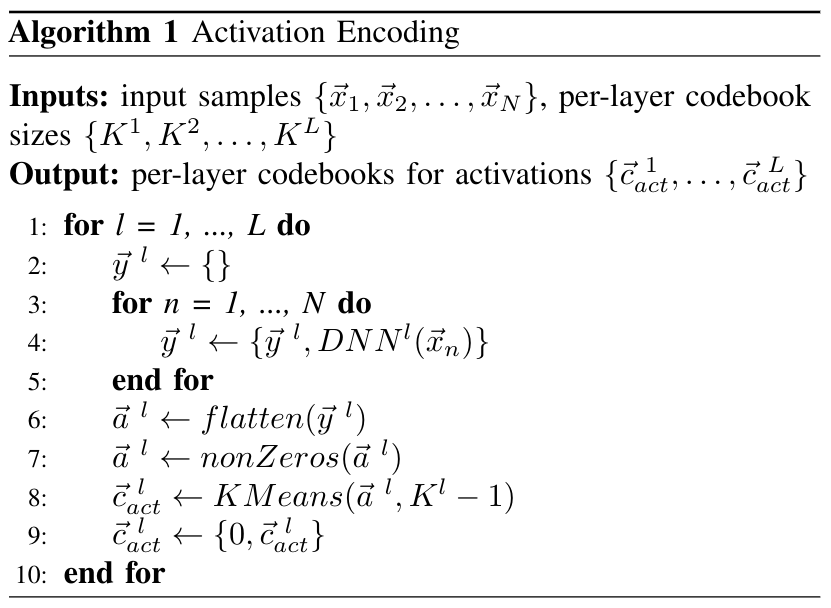}
    \vspace{-1em}
\end{figure}

\noindent{\bf Online Encoding.} The online encoding is performed during FPGA execution. The value of a feature $y$ is compared with the elements of the corresponding layer's codebook $\vec{c}_{act}^{\ l}$ to compute the encoding as $y_{enc}=argmin(|y-\vec{c}_{act}^{\ l}|)$.


\subsection{Training of Encoded Networks}\label{sec:training}
Encoding weights/parameters often results in a drop of accuracy. To compensate for such accuracy loss, the codebook entries are fine-tuned after encoding by means of a customized back-propagation scheme. In this section, we explain the details for fine-tuning encoded neural networks via Stochastic Gradient Descent (SGD)~\cite{kiefer1952stochastic}. For weights, the averaged gradient method~\cite{chen2015compressing, han2015deep} is applied. For activations, we develop new gradient computation methods as follows. 

\begin{figure}[h]
    \centering
    \begin{subfigure}[b]{0.4\columnwidth}
            \includegraphics[width=0.95\columnwidth]{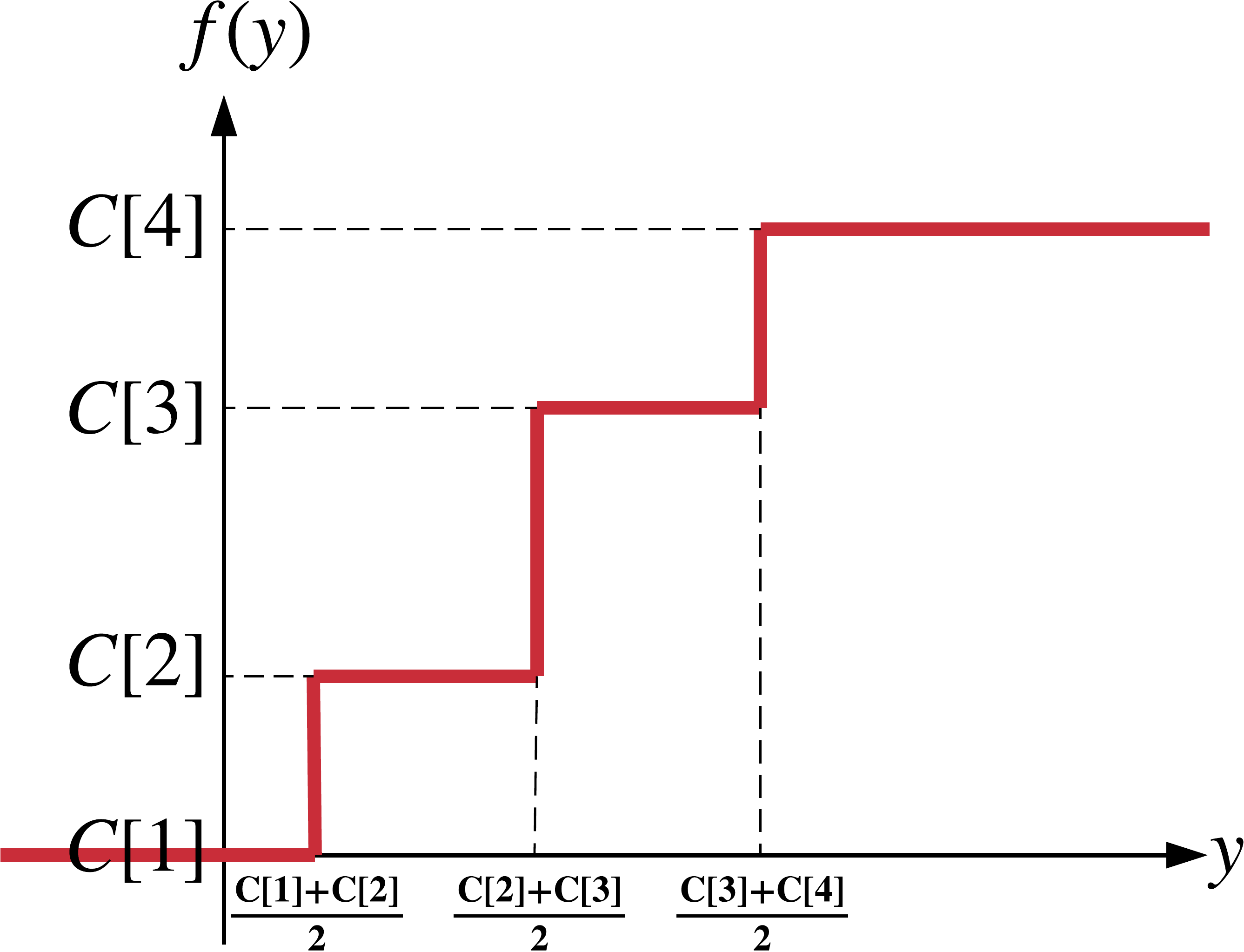}
            \caption{}
    \end{subfigure}
    ~
    \begin{subfigure}[b]{0.4\columnwidth}
        \includegraphics[width=0.95\columnwidth]{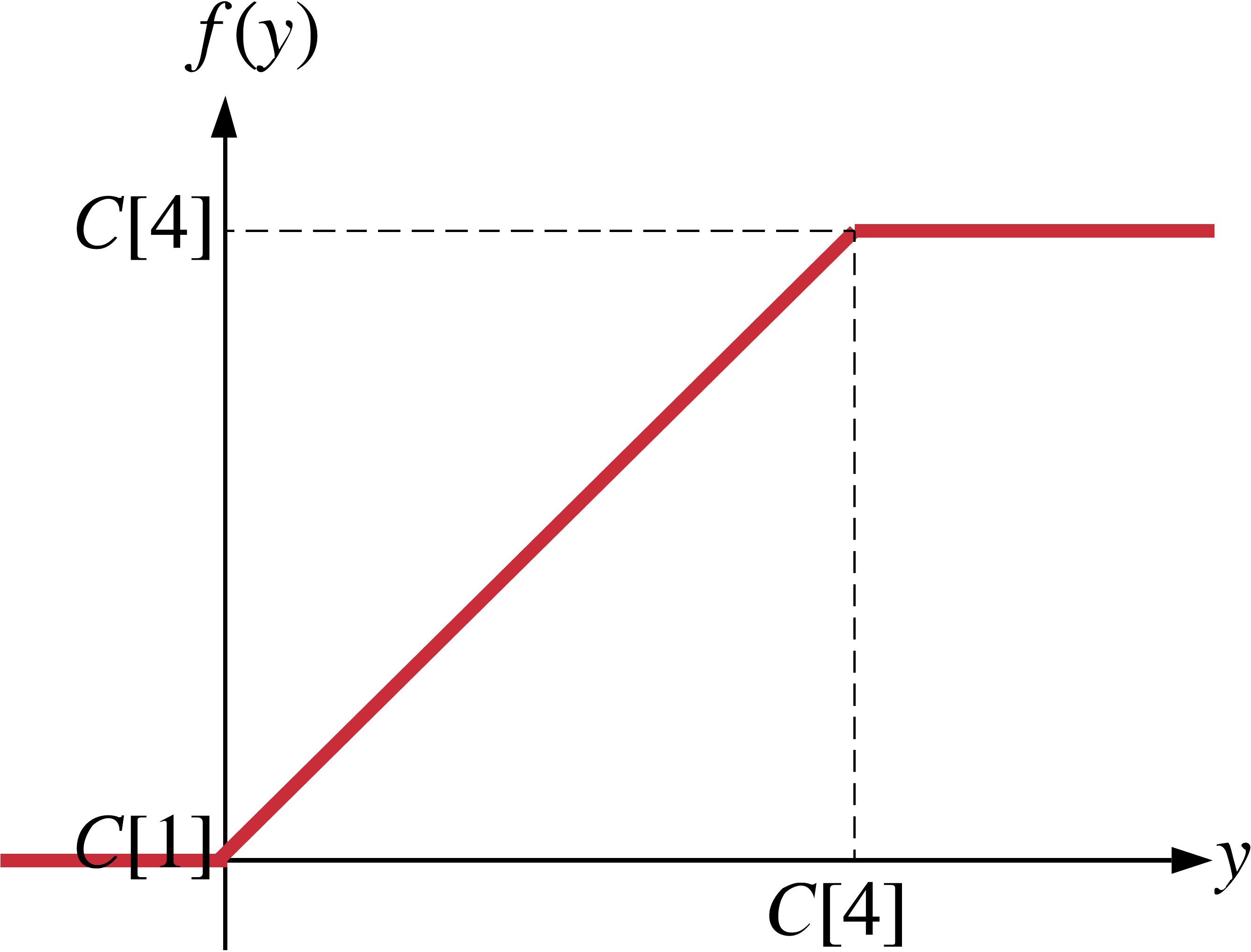}
        \caption{}
    \end{subfigure}
    \vspace{-0.5em}
    \caption{Example encoding nonlinearity with a codebook of ${K=4}$ elements. (a)~Nonlinear function applied in the forward propagation. (b)~Smooth approximation of encoding used in backward propagation for gradient computation.}
    \label{fig:enc_act_nonlinear}
    \vspace{-1.5em}
\end{figure}

Feature encoding can be viewed as a nonlinear transform which is made up of multiple \mystyle{step} functions as depicted in  Figure~\ref{fig:enc_act_nonlinear}. 
Given the gradient of the loss function with respect to the encoded values, $\nabla_{y^*}=\frac{\partial \mathcal{L}}{\partial y^*}$, we aim to compute the partial derivatives with respect to the non-encoded values ($\nabla_{y}=\frac{\partial \mathcal{L}}{\partial y}$) and the derivatives with respect to the codebook ($\vec{\nabla}_{c}=\frac{\partial \mathcal{L}}{\partial c}$). 



\noindent{\bf Computing $\mathbf{\nabla_{y}}$.} Given the partial derivative $\nabla_{y^*}$,  the gradient $\nabla_{y}$ can be obtained by applying the chain rule:
\begin{equation}
\resizebox{0.4\columnwidth}{!}{
    $\nabla_{y}=\frac{\partial \mathcal{L}}{\partial y}=\frac{\partial \mathcal{L}}{\partial y^*} \times \frac{\partial y^*}{\partial y}.$
}
\vspace{-0.2em}
\end{equation}

This formulation, however, is not stable due to the non-differentiability of the nonlinear function $f(\cdot)$. To address this issue, we propose to approximate the derivative of $f(\cdot)$ as:
\begin{equation}
\resizebox{0.55\columnwidth}{!}{
    $\frac{\partial f(y)}{\partial y}=\begin{cases} 
      1 & if\  c[1]<y<c[K] \\
      0 & otherwise 
   \end{cases},$
}
\end{equation}
where $c[1]$ and $c[K]$ are the smallest and largest codebook values, respectively. During forward propagation, $y^*$ is computed as shown in Figure~\ref{fig:enc_act_nonlinear}-a, whereas the backward propagation assumes the smooth function in Figure~\ref{fig:enc_act_nonlinear}-b. 

\noindent{\bf Computing $\mathbf{\nabla_{c}}$.} Given a scalar gradient element  $\nabla_{y^*}$, the gradient with respect to $c[k]$ is computed as:
\begin{equation}
\resizebox{0.4\columnwidth}{!}{
    $\vec{\nabla}_{c}[k]=I(c[k],y^{*})\times \nabla_{y^*},
    \vspace{-0.2em}$
}
\end{equation}
with $I(a,b)=1$ if $a=b$ and zero otherwise (identity operator).
Given a vector of features $\vec{y}^{*}$ and the corresponding vector of gradients $\vec{\nabla}_{{y}^{*}}$, the partial derivative is:
\begin{equation}
\resizebox{0.5\columnwidth}{!}{
    $\vec{\nabla}_{c}[k]=\sum_jI(c[k],\vec{y}^{*}[j])\times \vec{\nabla}_{y^*}[j].$
}
\vspace{-0.5em}
\end{equation}

Once we compute these partial derivatives, standard back-propagation algorithms can be used to fine-tune DNN parameters. We incorporate the gradient computation routines into \sys{} software stack to support fine-tuning for encoded DNNs. As we show in our evaluations, \sys{} fine-tuning phase has negligible runtime overhead compared to the original training. 

\subsection{Automated Bitwidth Customization}\label{sec:customization}
To alleviate design engineering cost, we propose a bitwidth customization algorithm that automatically determines the encoding bitwidth across all DNN layers. \sys{} configures the per-layer numerical precision by changing the codebook size $K$, which translates to the encoding bitwidth $\mli{ceil(Log_2[K])}$; a larger codebook leads to a more accurate approximation which comes at the cost of a higher memory footprint.  

\sys{} performs a diminishing search over possible bitwidth configurations to capture the trade-off between model accuracy and memory footprint. Algorithm 2 depicts the pseudo-code for our proposed method. Each step of the algorithm starts with an initial configuration of bitwidths, $\mli{config_{i}}=\{b_1, \dots,b_L\}$ where $b_l$ is the $l$-th layer bitwidth and $i$ is the iteration number. Possible \textit{next} configurations are evaluated by sweeping the bitwidth of $l$-th layer from 1 to $b_l-1$ while all other bitwidths are fixed. For a new configuration that sets the $l$-th layer's bitwidth to $b$, we compute memory reduction $\Delta M(l,b)$ and accuracy drop $\Delta A(l,b)$. Among all candidate configurations, we select the one that maximizes the reward function $\frac{\Delta M(l,b)}{\Delta A(l,b)}$. 

The selected candidate is then used as the starting configuration of layer encoding bitwidths in the next iteration. The output of Algorithm 2 is a set of bitwidth configurations, each generated in a certain iteration, that capture the tradeoff between accuracy and the memory footprint. We emphasize that the bitwidth configuration algorithm does not perform any re-training of the DNN in between the iterations. In addition, the inner loop performs accuracy evaluation on a very small subset of validation data (line 8 of Algorithm 2). Therefore, the runtime of \sys{} customization is drastically smaller than the original DNN training as we show in the evaluation section.

\begin{figure}[h]
    \centering
    \vspace{-1em}
    \includegraphics[width=0.8\columnwidth]{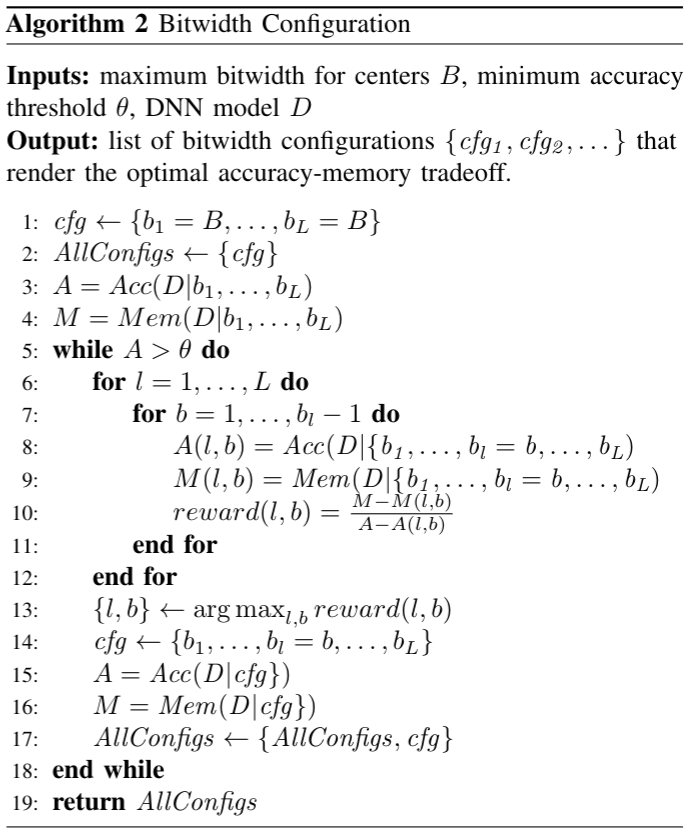}
    \vspace{-1.5em}
\end{figure}

\subsection{\sys{} Hardware Stack} \label{sec:hardware}
\sys{} is equipped with a hardware library described in high-level synthesis C++ programming language. This library consists of the essential building blocks to implement DNN layers (e.g, convolution, max-pooling, etc.) on FPGA. Each layer-type is implemented as a template function with certain computing engines that are customized to the specifications of the pertinent layer such as the input/output dimensions. By means of this tailoring, \sys{} fully exploits the benefits of FPGA reconfigurability and delivers a bit-flexible design. 

\begin{figure}[h]
    \centering
    \vspace{-1em}
    \includegraphics[width=0.65\columnwidth]{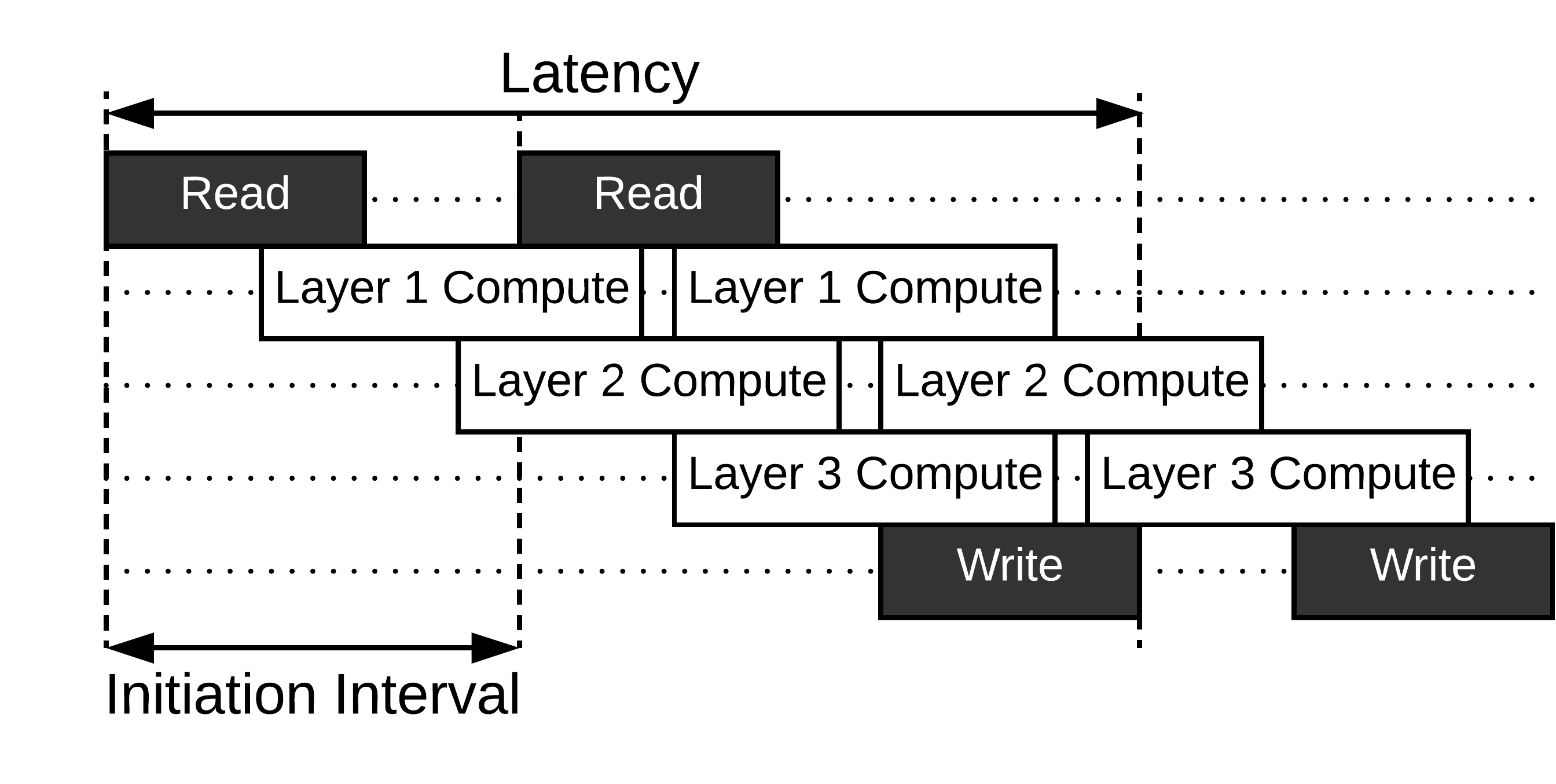}
    \vspace{-1em}
    \caption{Pipeline execution of layer computations in a streaming-based architecture increases the overall throughput. }
    \label{fig:overlay_latency}
\end{figure}

\sys{} adopts a streaming-based architecture which facilitates pipelining and overlays the computational overheads of subsequent DNN layers to increase the overall throughput and minimize latency. Figure~\ref{fig:overlay_latency} presents such pipelined execution for a DNN with $3$ layers. To ensure portability and efficiency, we chose the FINN~\cite{umuroglu2017finn} framework from Xilinx as the base of our hardware accelerator. FINN library was originally intended for execution of Binary Neural Networks (BNNs). We modified the original library to support operations on encoded parameters/weights and fixed-point MAC operations instead of the $\mli{XnorPopCount}$ operations required in BNNs~\cite{umuroglu2017finn}. 

\begin{figure}[h]
    \centering
    \vspace{-0.5em}
    \includegraphics[width=1\columnwidth]{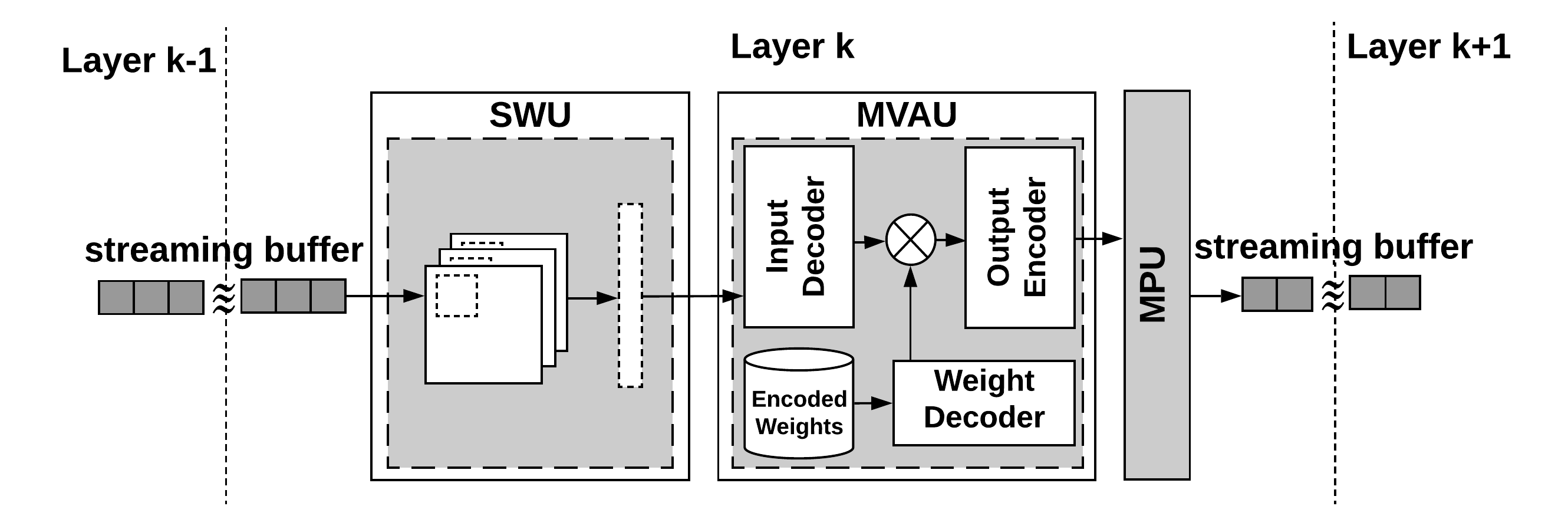}
    \caption{Schematic of \sys{} accelerator for encoded DNN inference. SWU reorders the input buffer, MVAU performs the core layer computations, and MPU implements max-pooling.}
    \label{fig:hardware_flow}
\end{figure}

Our accelerator is specifically designed to accommodate low-bitwidth encoded networks while supporting full-precision computations. Figure~\ref{fig:hardware_flow} depicts the flow diagram of \sys{} accelerator for implementing an encoded DNN on FPGA. The Sliding Window Unit (SWU) reorders the convolution layer input feature-maps to generate appropriate streaming buffers for the Matrix-Vector-Activation Unit (MVAU). The MVAU is the core computational module of \mystyle{CONV} and \mystyle{FC} layers which performs the matrix-vector multiplication, activation, and batch normalization. The Max-Pooling Unit (MPU) performs max-pooling over the feature-maps. In the following, we discuss the core modules \sys{} Hardware API.

\subsubsection{\textbf{Matrix-Vector-Activation Unit (MVAU)}}\label{sec:MVAU}
\begin{figure}[]
    \centering
    \vspace{-1em}
    \includegraphics[width=0.75\columnwidth]{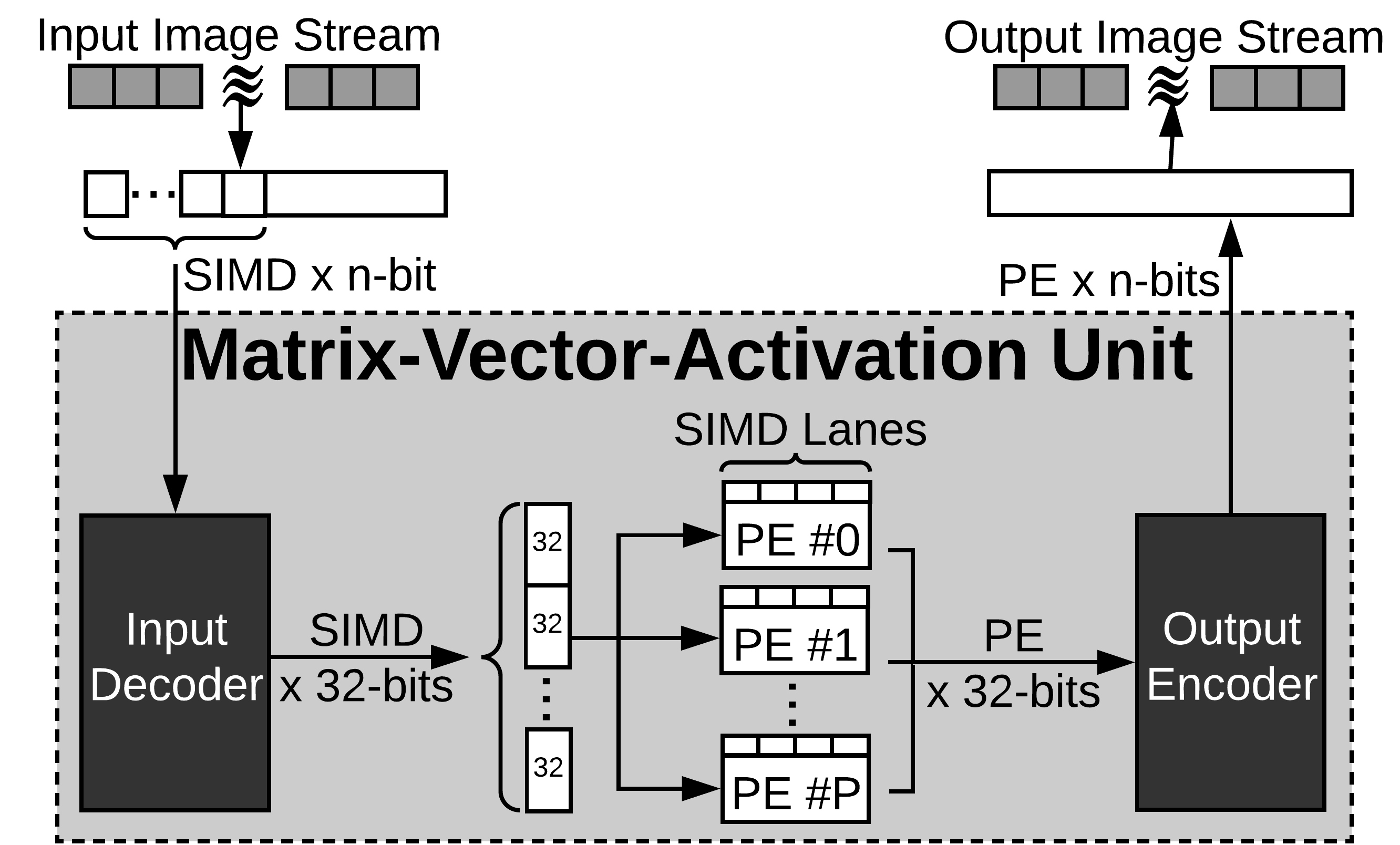}
    \caption{Computational flow of \sys{} MVAU. This unit performs matrix-vector multiplication, batch normalization, and a non-linear activation. The MVAU is equipped with an input decoder, an output encoder, and a weight decoder to comply with \sys{} encoded networks.} \label{fig:MVAU}
    \vspace{-1.5em}
\end{figure}

The MVAU in \sys{} hardware library is instantiated in both the convolution (\mystyle{CONV}) and fully-connected (\mystyle{FC}) layers to generate output features using the corresponding layer's specifications. Figure~\ref{fig:MVAU} illustrates the MVAU computational flow. This module performs $3$ fundamental tasks required in state-of-the-art DNNs, namely matrix-vector multiplication, batch normalization, and applying a non-linear activation. Internally, the MVAU is composed of an array of Processing Engines (\mystyle{PE}s) each of which accepts \mystyle{SIMD} lanes of input in parallel. In addition, \sys{} MVAU has customized encoding/decoding cores for output/input and layer weights.  

\noindent\textbf{Matrix-Vector Multiplication.} The main operations performed in linear DNN layers can be represented as a series of matrix-vector multiplications. 
The matrix-vector multiplication core in \sys{} MVAU offers two levels of parallelism to facilitate throughput control across DNN layers. Firstly, layer output generation is distributed among several \mystyle{PE}s working in parallel with each \mystyle{PE} responsible for generating the output of multiple feature-map channels (neurons) in a \mystyle{CONV} (\mystyle{FC}) layer. 
Furthermore, each \mystyle{PE} performs a number of Multiply-Accumulate (MAC) operations in parallel on \mystyle{SIMD} inputs. MAC operations are performed in fixed-point on decoded input/weight values, implemented using FPGA DSP slices.


\noindent\textbf{Decoder Modules.} Each layer in the encoded DNN contains two decoding codebooks corresponding to the inputs and weights. All codebook values are stored in fixed-point (e.g, 32 bits) and incur a negligible memory footprint. The decoder modules use the encoded values to address a small memory block storing the codebooks. 
To achieve maximum efficiency, the input decoder is designed such that the values are processed in groups of SIMD elements. To facilitate parallelism, each \mystyle{PE} owns a copy of the corresponding weight codebook. Upon execution of the multiply-accumulate operations, the replicated codebooks can be accessed in parallel to decode the weights.


\noindent\textbf{Encoder Module.} The combination of \mystyle{ReLU} activation function and output encoding is realized by the encoder module. This unit performs a nearest-neighbor search by comparing a fixed-point input value against the encoding codebook and transferring the index of the closest element through the output streaming buffer. Setting the first codebook value to $0$ ensures that the \mystyle{ReLU} functionality is inherently applied. 

\noindent\textbf{Batch Normalization.} For each neuron $x$ in the layer feature-map, batch normalization is equivalent to $\Tilde{x}=\alpha x+\beta$ where $\alpha$ is the scaling factor and $\beta$ is the bias value. These values are extracted by \sys{} compiler from the trained encoded DNN. In order to implement the batch normalization, each \mystyle{PE} in the MVAU performs a single multiplication (by value $\alpha$) and a single addition (with value $\beta$). 


\subsubsection{\textbf{Sliding Window Unit (SWU)}}\label{sec:SWU} 
The convolutional layers of a DNN compute the dot product between a window of the layer input and the \mystyle{CONV} weight kernel. The window is slid over the input image to produce individual elements of the output feature-map. The SWU in \sys{} hardware simulates the sliding window operation by reordering the values in the layer input image buffer. The input image values are then grouped in chunks of SIMD words to be sent to the MVAU sequentially for processing. 


\subsubsection{\textbf{Max-pooling Unit (MPU)}} \label{sec:MPU} 
\sys{} software stack outputs a sorted list of codebook values for the encoded layer activations where the clusters centers with higher values are mapped to larger cluster indices. 
This sorting is particularly useful since comparison over encoded values becomes equivalent to comparison over the original fixed-point values; therefore, \sys{} performs the max-pooling operation on low-bitwidth encoded values rather than the full-precision cluster centers. This approach provides two benefits: 1)~the memory overhead of the buffers in the MPU is considerably reduced. 2)~The logic cost of comparison between low-bitwidth encoded values is significantly smaller than the full-precision counterpart. 


\section{Experiments}\label{sec:experiments}



To evaluate \sys{} effectiveness, we perform proof-of-concept experiments on four different classification benchmarks, namely, MNIST, SVHN, CIFAR-10, and ImageNet. 
We implement \sys{} software API in Pytorch library. The hardware API is realized in Vivado\_HLS design suite. All hardware resource utilizations are gathered after performing place-and-route via Vivado Design Suite 2017.2. Throughput values are reported from  Vivado\_HLS 2017.2.

\subsection{\sys{} Automated Bitwidth Selection}

We illustrate our bitwidth selection algorithm (Section~\ref{sec:customization}) using the \mystyle{AlexNet} architecture~\cite{alexnetpaper} trained on ImageNet dataset which contains $50000$ test samples. To perform customization, we split these images into $49000$ test and $1000$ validation samples. The validation data is used for bitwidth selection whereas the test data is utilized to report classification accuracy.

\begin{figure}[h]
    \centering
    \includegraphics[width=1\columnwidth]{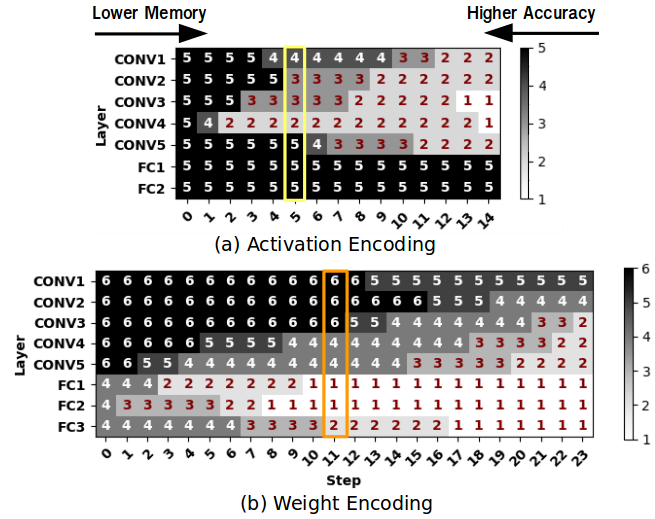}
    \caption{Per-layer bitwidth configuration at each step of \sys{} bitwidth selection algorithm, shown for AlexNet.
    } 
    \label{fig:heatmap}
\end{figure}

\vspace{0.2em}
\noindent{\textbf{Configuring Activation Bitwidths.}} The first step of \sys{} bitwidth customization is to encode the activations while the weights are kept at full-precision. Figure~\ref{fig:heatmap}-a demonstrates the iterative process of bitwidth selection for network activations. Here, each colored square at location $(l, i)$ represents the encoding bitwidth for the $l$-th layer's activation at step $i$ of the algorithm where $l$ and $i$ span the vertical and horizontal axes, respectively. Initially, the activations are encoded with $b_{enc}=5$ bits. As the algorithm proceeds, both the total memory footprint and DNN accuracy are decreased. The algorithm terminates when the accuracy drops below a threshold of $10\%$. We next select one of the configurations (corresponding to the column with bold border) in Figure~\ref{fig:heatmap}-a and perform fine-tuning to recover accuracy. We then proceed to the weight encoding step.

\vspace{0.2em}
\noindent{\textbf{Configuring Weight Bitwidths.}} Initially, the \mystyle{CONV} and \mystyle{FC} weights are encoded with $b_{enc}=6$ and $b_{enc}=4$ bits, respectively. During this customization step, the activation bitwidths remain unchanged and only the weight bitwidths are configured. The corresponding bitwidths of network weights in different iterations of the algorithm are depicted in Figure~\ref{fig:heatmap}-b and the selected weight configuration is highlighted by bold borders. Note that \sys{} enjoys the flexibility of selecting various pairs of activation-weight configurations depending on accuracy and memory constraints.

\vspace{0.2em}
\noindent{\textbf{Discussion on Selected Bitwidths.}} \sys{} aims at capturing the memory/accuracy tradeoff spanned by the bitwidth configurations. Consider the selected configuration of Figure~\ref{fig:heatmap}-b as an example. \mystyle{CONV} layers have more effect on model accuracy, requiring a high encoding bitwidth. In addition, the contribution of \mystyle{FC} layers to the overall weight memory footprint is much higher than that of \mystyle{CONV} layers. As a result, \mystyle{FC} layers are encoded with fewer bits compared to \mystyle{CONV} layers.



\begin{figure}[h]
    \centering
    \includegraphics[width=1\columnwidth]{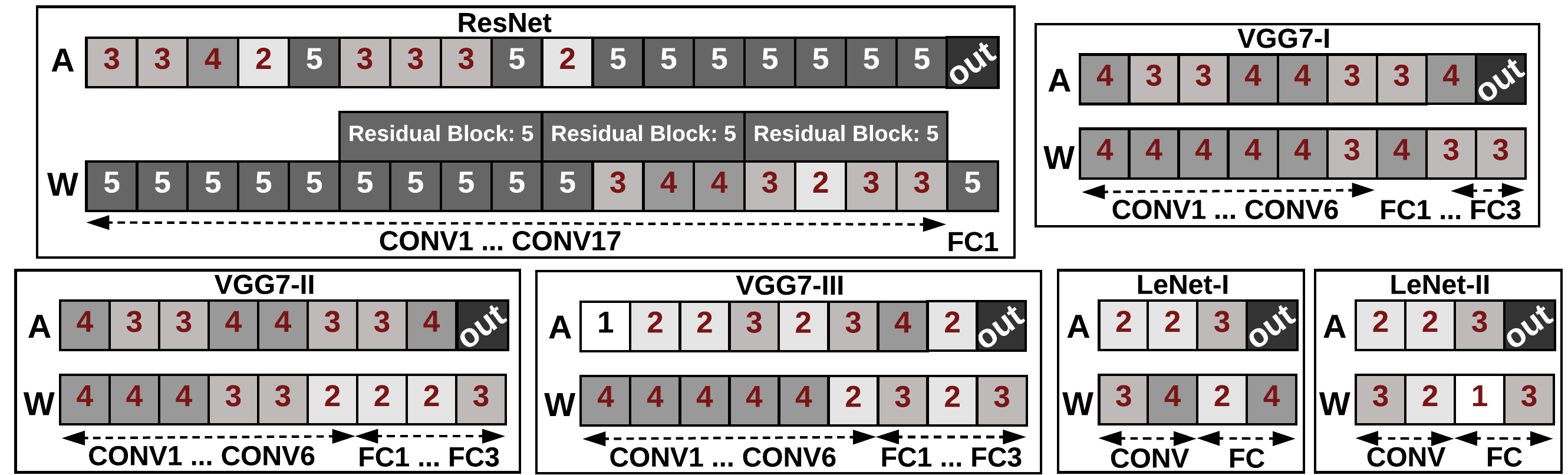}
    \caption{Per-layer encoding bitwidths for evaluated DNNs.
    }
    \label{fig:bitwidths_across}
\end{figure}

\vspace{0.2em}
\noindent{\textbf{Comparison with Prior Art.}} To compare \sys{} with existing low-bit neural networks, we train several widely popular architectures summarized in Table~\ref{tab:net_sum}. We then select a set of customized cross-layer encoding configurations generated by \sys{} bitwidth selection mechanism. Each of our examined architectures and their corresponding bitwidths are chosen specifically to match the accuracy and/or memory footprint of prior art. As such, for some architectures, e.g, \mystyle{VGG7}, multiple configurations are evaluated that merely differ in number of weight bits. Figure~\ref{fig:bitwidths_across} shows the selected per-layer bitwidths for the weights and activations of each DNN configuration. 

\begin{table}[h]
\centering
\caption{Summary of evaluated network architectures in terms of the number of convolutions (\mystyle{CONV}), max-pooling (\mystyle{MP}), global average pooling (\mystyle{AP}), and fully connected (\mystyle{FC}) layers.}\label{tab:net_sum}
\resizebox{0.75\columnwidth}{!}{
\begin{tabular}{ccc}
Dataset                                                      & Network                       & Description                                                 \\ \hline
MNIST                                                        & LeNet                         & 2 \mystyle{CONV}, 2 \mystyle{MP}, 2 \mystyle{FC} \\ \hline
\begin{tabular}[c]{@{}c@{}}SVHN\\ \&\\ CIFAR-10\end{tabular} & VGG7                          & 6 \mystyle{CONV}, 2 \mystyle{MP}, 3 \mystyle{FC} \\ \hline
\multirow{2}{*}{ImageNet}                                    & AlexNet                       & 5 \mystyle{CONV}, 3 \mystyle{MP}, 3 \mystyle{FC} \\
                                                             & \multicolumn{1}{l}{ResNet} & \multicolumn{1}{l}{17 \mystyle{CONV}, 1 \mystyle{MP}, 1 \mystyle{AP}, 1 \mystyle{FC} }    \\ \hline 
\end{tabular}
}
\end{table}

Table~\ref{tab:acc_and_mem_comp} compares the selected configurations with prior art in terms of memory, accuracy, and fine-tuning time. It is worth mentioning that, unlike existing low-bit DNNs which train the whole network from scratch, \sys{} takes a post-processing approach that is readily applied to pre-trained neural networks. The importance of this property is two-fold: (i)~\sys{} eliminates the drastic cost of training from scratch per bitwidth configuration. Using our fine-tuning method explained in Section~\ref{sec:training}, the model accuracy is retrieved after few epochs, e.g., as low as 0.25 epochs for ImageNet. (ii)~\sys{} customization can be applied to pre-trained models that are publicly available online. 

\begin{table}[h]
\caption{Comparison of \sys{} with state-of-the-art low-bit DNNs. The per-layer bitwidths for AlexNet are shown in Figure~\ref{fig:heatmap}. The bitwidths for LeNet-X, VGG7-X, and ResNet models are same as those in Figure~\ref{fig:bitwidths_across}.}\label{tab:acc_and_mem_comp}
\resizebox{1\columnwidth}{!}{
\begin{tabular}{p{0.7cm}p{1.7cm}p{1.2cm}p{0.6cm}p{0.6cm}p{1cm}p{0.6cm}p{0.6cm}}
\cline{2-8}
        & \multirow{2}{*}{Baselines} & \multirow{2}{*}{Arch}     & \multirow{2}{*}{\begin{tabular}[c]{@{}c@{}}Test \\ Acc\end{tabular}} & \multirow{2}{*}{Mem} & \multirow{2}{*}{Epochs} & \multicolumn{2}{c}{Bitwidth}                         \\ \cline{7-8}
        \multicolumn{1}{l}{}                            &                            &                           &                                                                           &                      &                         & \multicolumn{1}{l}{Weight} & \multicolumn{1}{l}{Act} \\ \hline
\multicolumn{1}{c|}{\multirow{4}{*}{MNIST}}     & QNN~\cite{hubara2016quantized}         & MLP                       & 99.04                                                    & $193\times$  & 1000      & ~~~1      & ~~1   \\
\multicolumn{1}{c|}{}                           & ReBNet3~\cite{ghasemzadeh2018rebnet}     & MLP                       & 98.25                                                    & $1.76\times$ & 200       & ~~~1      & ~~2   \\
\multicolumn{1}{c|}{}                           & \sys{}   & LeNet-I                   & 99.02                                                    & $1.84\times$ & $10\times2^*$ & \multicolumn{2}{c}{flexible}   \\
\multicolumn{1}{c|}{}                           & \sys{}   & LeNet-II                  & 98.31                                                    & $1$          & $10\times2$     & \multicolumn{2}{c}{flexible}   \\ \hline
\multicolumn{1}{c|}{\multirow{4}{*}{CIFAR-10}}  & QNN~\cite{hubara2016quantized}         & VGG7                      & 89.85                                                    & $5.4\times$  & 500       & ~~~1      & ~~1   \\
\multicolumn{1}{c|}{}                           & ReBNet3~\cite{ghasemzadeh2018rebnet}     & VGG7                      & 86.98                                                    & $1.65\times$ & 200       & ~~~1      & ~~3   \\
\multicolumn{1}{c|}{}                           & \sys{}   & VGG7-I                    & 88.14                                                    & $1.32\times$ & $10\times2$     & \multicolumn{2}{c}{flexible}   \\
\multicolumn{1}{c|}{}                           & \sys{}   & VGG7-II                   & 87.01                                                    & $1$          & $10\times2$     & \multicolumn{2}{c}{flexible}   \\ \hline
\multicolumn{1}{c|}{\multirow{3}{*}{SVHN}}      & ReBNet3~\cite{ghasemzadeh2018rebnet}     & VGG7                      & 97.00                                                    & $1.62\times$ & 50        & ~~~1      & ~~3   \\
\multicolumn{1}{c|}{}                           & QNN~\cite{hubara2016quantized}         & VGG7                      & 97.2                                                     & $2.15\times$ & 200       & ~~~1      & ~~1   \\
\multicolumn{1}{c|}{}                           & \sys{}   & VGG7-III                  & 97.15                                                    & $1$          & $10\times2$     & \multicolumn{2}{c}{flexible}   \\ \hline
\multicolumn{1}{c|}{\multirow{10}{*}{ImageNet}} & HWGQ~\cite{cai2017deep}        & \multirow{7}{*}{AlexNet}  & 52.70                                                    & $1.17\times$ & 68        & ~~~1      & ~~2   \\
\multicolumn{1}{c|}{}                           & QNN~\cite{hubara2016quantized}         &                           & 51.03                                                    & $1.17\times$ & -         & ~~~1      & ~~2   \\
\multicolumn{1}{c|}{}                           & DoReFaNet~\cite{zhou2016dorefa}   &                           & 49.80                                                    & $1.17\times$ & 45        & ~~~1      & ~~2   \\
\multicolumn{1}{c|}{}                           & WRPN~\cite{mishra2017wrpn}$^{\ddagger}$ &                           & 48.30                                                    & $4.61\times$ & -         & ~~~1      & ~~1   \\
\multicolumn{1}{c|}{}                           & XNORNet~\cite{rastegari2016xnor}     &                           & 44.20                                                    & $1.15\times$ & 18        & ~~~1      & ~~1   \\
\multicolumn{1}{c|}{}                           & ReBNet3~\cite{ghasemzadeh2018rebnet}     &                           & 41.43                                                    & $1.17\times$ & 100       & ~~~1      & ~~3   \\
\multicolumn{1}{c|}{}                           & \sys{}   &                           & 53.21                                                    & $1$          & $0.25\times2$ & \multicolumn{2}{c}{flexible}   \\ \cline{2-8} 
\multicolumn{1}{c|}{}                           & ABC-Net~\cite{lin2017towards}     & \multirow{3}{*}{ResNet18} & 65.00                                                    & $1.57\times$ & -         & ~~~5      & ~~5   \\
\multicolumn{1}{c|}{}                           & ABC-Net~\cite{lin2017towards}     &                           & 62.50                                                    & $1.05\times$ & -         & ~~~3      & ~~5   \\
\multicolumn{1}{c|}{}                           & \sys{}   &                           & 65.40                                                    & $1$          & $0.25\times2$ & \multicolumn{2}{c}{flexible}   \\ \hline
\vspace{0.1cm}
\end{tabular}}
\scriptsize{$^*$fine-tuning for 10 epochs post-activation and 10 epochs post-weight encoding.\\
$^\ddagger$ This baseline has $2\times$ more neurons per layer.}
\end{table}

\noindent{\bf Overhead of Customization and Re-training:}
Table~\ref{tab:overhead} report \sys{} customization overhead for our most complex benchmark, i.e., ResNet. The customization phase incurs roughly $8.5\%$ of the training time. Note that the bitwidth customization phase is performed offline, prior to design synthesis and hardware evaluation. 


\begin{table}[h]
\centering
\caption{Breakdown of customization overhead for ResNet-18 network, namely, activation encoding (AE), fine-tuning (FT), and weight encoding (WE). The number of training epochs is extracted from the original ResNet paper~\cite{he2016deep}. The timing is reported for a machine running a single Nvidia Titan Xp GPU.}\label{tab:overhead}
\resizebox{1\columnwidth}{!}{
\begin{tabular}{cccccc}
\hline
\multirow{2}{*}{Customization} & AE       & FT     & WE      & FT     & Total      \\ \cline{2-6} 
                               & 114 mins & 4 mins & 31 mins & 4 mins & \textbf{153 mins}   \\ \hline
Training                       & -        & -      & -       & -      & \textbf{1800  mins} \\ \hline
\end{tabular}}
\end{table}


\subsection{\sys{} Hardware Implementation}
In this section, we evaluate \sys{} hardware accelerator. We implement one architecture per dataset from Figures~\ref{fig:heatmap} and~\ref{fig:bitwidths_across}, namely, LeNet-I for MNIST, VGG7-I for CIFAR10, VGG7-III for SVHN, and AlexNet for ImageNet. Table~\ref{tab:platforms} summarizes the evaluation platforms for each DNN architecture. 

\begin{table}[h]
\centering
\caption{Platform details in terms of block ram (BRAM), DSP, flip-flop (FF), and look-up table (LUT) resources.}
\label{tab:platforms}
\resizebox{\columnwidth}{!}{
\begin{tabular}{llllll}
\hline
Application      & Platform       & BRAM & DSP & FF      & LUT    \\ \hline
ImageNet         & Virtex VCU108  & 3456 & 768 & 1075200 & 537600 \\
CIFAR-10 \& SVHN & Zynq ZC702     & 280  & 220 & 106400  & 53200  \\
MNIST            & Spartan XC7S50 & 120  & 150 & 65200   & 32600  \\ \hline
\end{tabular}
}
\end{table}

\vspace{0.2em}
\noindent\textbf{Importance of Activation Encoding.}
We start the analysis by studying the advantages of activation encoding, from the hardware perspective, versus solely encoding the weights as proposed in~\cite{han2015deep}. Note that~\cite{han2015deep} also applies pruning and Huffman encoding which are the main contributors to the compression rate. Since these methods are orthogonal to our approach, we do not utilize them in \sys{} to focus on the analysis of encoding itself. We compare two versions of encoded DNNs: one with encoded weights and fixed-point activations as proposed in~\cite{han2015deep}, and another with both weights and activations encoded as suggested in \sys{}. 
For each dataset, we separately optimize the per-layer parallelism factors \mystyle{SIMD} and \mystyle{PE} for both encoded and fixed-point DNNs to obtain maximum throughput while complying with the resource constraints. Table~\ref{tab:fpga_implementation_results} summarizes the resource utilization and throughput for each of the designs. 

Overall, realization of \sys{} methodology, i.e., $\mli{DNN_{enc}}$, achieves higher throughput while requiring lower number of resources compared to a weight-only encoding approach~\cite{han2015deep}. The benefits become more prominent for architectures with higher complexity since the memory implication of activations contributes more in complex networks. As seen for MNIST, CIFAR-10, and SVHN benchmarks, \sys{} activation encoding improves the throughput by $1.1\times$, $6.2\times$, and $6.66\times$, respectively. The effect of activation encoding is most significant for the ImageNet benchmark; the memory of the model with fixed-point activations is larger than the capacity of the on-chip streaming buffers, rendering the design infeasible within platform constraints. For this dataset, we compare $\mli{DNN_{enc}}$ with existing fixed-point accelerators in the following.

\begin{table}[h]
\caption{Summary of hardware resource utilization and performance. $\mli{DNN_{enc}}$ stands for the model with encoded weights and activations whereas $\mli{DNN_{fix}}$ denotes the network with encoded weights and (8-bit) Fixed-point activations.
}\label{tab:fpga_implementation_results}
\resizebox{\columnwidth}{!}{
\begin{tabular}{ccccccc}
\cline{3-7}
                          &          & \multicolumn{4}{c}{Resource Utilization}                  & \multirow{2}{*}{\begin{tabular}[c]{@{}c@{}}Latency\\ (ms)\end{tabular}} \\ \cline{3-6}
                          &          & BRAM         & DSP$^*$          & FF           & LUT          &                                                                         \\ \hline
\multirow{3}{*}{MNIST}    & $\mli{DNN_{enc}}$      & 33           & 53           & 15223        & 9992         & 0.39                                                                    \\
                          & $\mli{DNN_{fix}}$      & 93           & 53           & 25884        & 12048        & 0.43                                                                    \\
                          & $\mli{Ratio}$ & $2.82\times$ & $1\times$    & $1.7\times$  & $1.2\times$  & \textbf{1.1}$\times$                                                             \\ \hline
\multirow{3}{*}{CIFAR-10} & $\mli{DNN_{enc}}$      & 197          & 111          & 53953        & 31632        & 3.58                                                                    \\
                          & $\mli{DNN_{fix}}$      & 181          & 35           & 68255        & 28433        & 22.21                                                                   \\
                          & $\mli{Ratio}$ & $0.92\times$ & $0.32\times$ & $1.26\times$ & $0.9\times$  & \textbf{6.2}$\times$                                                             \\ \hline
\multirow{3}{*}{SVHN}     & $\mli{DNN_{enc}}$      & 146          & 111          & 42748        & 28393        & 3.39                                                                    \\
                          & $\mli{DNN_{fix}}$      & 143          & 35           & 67944        & 27934        & 22.59                                                                   \\
                          & $\mli{Ratio}$ & $0.98\times$ & $0.32\times$ & $1.59\times$ & $0.98\times$ & \textbf{6.66}$\times$                                                       \\ \hline
\multirow{2}{*}{ImageNet} & $\mli{DNN_{enc}}$      & 3336         & 308          & 159663       & 82791        & 25.05                                                                   \\
                          & $\mli{DNN_{fix}}$      & \multicolumn{5}{c}{Exceeds Platform Constraints}                                                                    \\ \hline
\end{tabular}
}
\scriptsize{$^*25\times18$  DSB array.}
\vspace{-0.5em}
\end{table}

\noindent{\bf Comparison with Fixed-point Accelerators.} We perform a comprehensive comparison between \sys{} and prior work in Table~\ref{tab:chaidnn_compare}. Specifically, we consider AlexNet with customized encoding as in Figure~\ref{fig:heatmap}, which corresponds to hardware results of Table~\ref{tab:fpga_implementation_results}. The reported results include performance, either in terms of throughput (frames per second) or latency. Since the existing frameworks utilize various hardware platforms, it is crucial to take into account the instantiated computational capacity\footnote{the computational capacity is defined as $\mli{CAP=DSP\times Arr}$,  where $Arr$ is the array size per DSP, e.g., $18\times25$ for Xilinx Virtex platforms.} and power consumption. Therefore, we compare the frameworks by means of performance-per-resource and performance-per-Watt.
\sys{} achieves higher normalized performance compared to prior art. This is a direct result of using on-chip memory instead of the off-chip DRAM for feature transfer among DNN layers. The streaming buffers of our design allow \sys{} to better utilize the arithmetic units by overlapping the execution of DNN layers, achieving a higher performance-per-resource. The power advantage of \sys{} over existing accelerators is also rooted in the elimination of power-hungry DRAM access.



%

\begin{table}[h]
\caption{Comparison of Alexnet implementation between \sys{} and existing fixed-point (FXD) and floating-point (FLT) DNN accelerators. To account for platform variations, we compare the throughput (frames-per-second) and $\frac{\textrm 1}{\textrm {Latency}}$ metrics normalized by computation capacity (CAP). We also compare performance-per-Watt to reflect power efficiency. }\label{tab:chaidnn_compare}
\resizebox{1\columnwidth}{!}{
\begin{tabular}{p{1.1cm}p{0.95cm}p{0.6cm}p{0.6cm}p{0.6cm}p{0.6cm}p{0.6cm}p{0.6cm}p{0.85cm}}
\hline                                                                 & \multicolumn{1}{c}{Criterion} &  \cite{shen2017maximizing}  & \cite{suda2016throughput}  & \cite{zhang2016energy}     & \cite{liu2017throughput}           & \cite{sharma2016high} &  \cite{chaidnn}      & \sys{}             \\ \hline
                                                                            & Precision                     & FLT          & FXD           & FXD             & FXD             & FXD                              & FXD          & Flexible       \\
                                                                            & Acc(\%)                & ~~-        & 55.41         & 52.4            & 56.5            & ~~-                                & 54.27        & 53.2           \\
                                                                            & FPGA                           & 690T$^*$         & GSD8$^\dagger$         & 690T$^*$         & 690T$^*$        & AX115$^\ddagger$                            & ZU9$^{\S}$        & VCU108$^{*}$ \\
                                                                            & Freq(MHz)                    & 100          & 120           & 150             & 100             & 200                              & 300          & 152            \\
                                                                            & DSP$^{**}$                           & 3177         & 1504          & 14400           & 2872            & 2688                             & 442          & 308            \\ \hline
\multirow{2}{*}{Throughput}  & /CAP                      & 0.55$\times$ & 1  & 0.88$\times$    & 2.33$\times$    & 1.42$\times$                     & 0.49$\times$ & \textbf{3.03}$\times$              \\
                                                                            & /Watt                     & 3.14$\times$ & 1  & 1.82$\times$    & \textbf{5.00}$\times$    & 1.36$\times$                     & ~~~-            & 4.54$\times$              \\ \hline
\multirow{2}{*}{\large $\frac{\textrm 1}{\textrm {Latency}}$}                                                  & /CAP                      & ~~~-            & 1 & 0.05$\times$   & 0.15$\times$   & ~~~-                                & ~~~-            & \textbf{3.03}$\times$              \\
                                                                            & /Watt                     & ~~~-            & 1  & 0.10$\times$   & 0.32$\times$   & ~~~-                                & ~~~-            & \textbf{4.54}$\times$             \\ \hline
\end{tabular}
}
\scriptsize{$^*$Virtex\ \ \ \              $^\dagger$Stratix-V\ \ \ \ $^\ddagger$Arria10\ \ \ \ $^{\S}$Zynq\\
$^{**}$DSP array size is $25\times 18$ for Xilinx and $18\times18$ for Altera/Intel FPGAs.}

\end{table}

\begin{figure}[h]
    \centering
    \includegraphics[width=0.7\columnwidth]{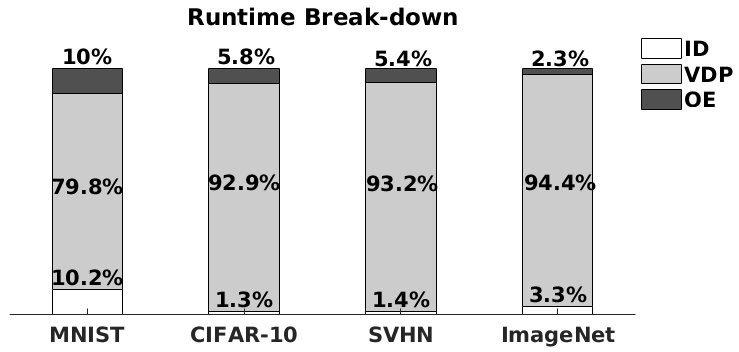}
    \caption{Runtime breakdown of Input Decoding (\mystyle{ID}), Vector-Dot-Product (\mystyle{VDP}), and Output Encoding (\mystyle{OE}) across layers.}
    \vspace{-0.5em}
    \label{fig:encoding_overhead}
\end{figure}

\noindent{\bf Execution Overhead of Encoding/Decoding.} We study the runtime implication of online activation encoding by measuring the number of clock cycles required for different stages of \sys{} \mystyle{MVAU} engine. Figure~\ref{fig:encoding_overhead} demonstrates the runtime break-down for each of the evaluated architectures. For a conventional non-encoded network, the \mystyle{MVAU} would only perform Vector-Dot-Product (\mystyle{VDP}) operations. As can be seen, for \sys{} encoded models, the majority of clock cycles in MVAU execution belong to \mystyle{VDP} computation while the encoding/decoding overhead is negligible.

\section{Related Work}\label{sec:related}
As neural networks are memory-intensive, devising methods to decrease the memory footprint can significantly enhance accelerator performance in terms of throughput and power consumption. An attractive solution for memory reduction is training few-bit DNNs. Several methods for training DNNs with few bits have been proposed in~\cite{zhou2016dorefa,mishra2017wrpn,cai2017deep,rastegari2016xnor,hubara2016quantized,lin2017towards}. These papers focus on the establishment of the theoretical foundation for low-bit DNN inference. Nevertheless, the inference accuracy of these DNNs is often lower than the original full-precision model. Higher accuracy can be achieved using DNN models that perform computations in the fixed-point regime. 

Nonlinear encoding allows for utilization of fixed-point arithmetics accompanied by a low storage requirement. Perhaps the closest method to this paper is a stand-alone weight encoding, with no activation encoding, originally proposed in~\cite{han2015deep, chen2015compressing, samragh2017customizing}. Weight encoding significantly reduces the memory footprint of model parameters but the activation units (especially in convolution layers) still require a large capacity of memory. To address this challenge, we extend the encoding to the activations of neural networks and introduce training routines for the corresponding encoded activations. In addition, prior work utilizes hand-crafted or rule-based heuristics to determine the encoding bitwidth. Such manual methods are generally sub-optimal and incur a drastic engineering cost. To address this issue, we propose an automated cross-layer bitwidth selection algorithm that aims to capture the accuracy/memory tradeoff.

In a concurrent track, designing automated and easy-to-use tools for FPGA implementation of DNNs has been the focus of contemporary research~\cite{shen2017maximizing,suda2016throughput,zhang2016energy,liu2017throughput,sharma2016high,chaidnn}. These works aim to maximize the throughput of fixed/floating-point DNN inference by distributing FPGA resources among parallel computational engines. Although accurate, fixed-point DNNs are generally memory intensive, where excessive access to off-chip memory becomes a design bottleneck. To alleviate this problem, authors of~\cite{umuroglu2017finn,ghasemzadeh2018rebnet} propose to perform inference solely using the on-chip memory and utilizing streaming buffers to realize inter-layer data transfers. These frameworks facilitate the design process of DNNs by providing configurable template functions in high-level synthesis language. However, ~\cite{umuroglu2017finn,ghasemzadeh2018rebnet} are only compatible with binary DNNs as their core computational engines do not support fixed-point arithmetics. By incorporating activation encoding into DNN computational flow, \sys{} hardware simultaneously enjoys the performance benefits of on-chip streaming buffers and the high accuracy of fixed-point arithmetics. \sys{} hardware stack supports flexible bitwidths, allowing the implementation of customized encoded DNNs.
\section{Conclusion}
This paper proposes a novel nonlinear quantization scheme to reduce the memory footprint of intermediate activations in convolutional neural networks computation flow. The encoding compresses the activations and allows on-chip execution of the underlying FPGA accelerator without communicating the computed features with the off-chip DRAM. To ensure non-recurring engineering cost, an automated algorithm is proposed to configure the encoding bitwidth across all layers of an arbitrary neural network. The open-source API of \sys{} enables developers to convert high-level Pytorch description of the neural network into hardware description without getting involved with the details of the design. We hope the provided API can advance research on reconfigurable DNN inference.


\bibliographystyle{IEEEtran}
\bibliography{ref}

\end{document}